# An Integrated System of Drug Matching and Abnormal Approval Number Correction

Chenxi Dong
*School of Data Science*
City University of Hong Kong
Hong Kong, China
chenxdong3-c@my.cityu.edu.hk

Junchang Zhang
*School of Data Science*
City University of Hong Kong
Hong Kong, China
richard.zhang@my.cityu.edu.hk

Dalue Lin
*School of Data Science*
City University of Hong Kong
Hong Kong, China
daluelin2-c@my.cityu.edu.hk

Qingpeng Zhang
*School of Data Science*
City University of Hong Kong
Hong Kong, China
qingpeng.zhang@cityu.edu.hk

Zhichao Yan
111, Inc.
Shanghai, China
yanzhichao@111.com.cn

Yuechun Yu
111, Inc.
Shanghai, China
yuyuechun@111.com.cn

Cao Kang
111, Inc.
Shanghai, China
kangcao@111.com.cn

Bin Hu*
111, Inc.
Shanghai, China
hubin@111.com.cn

*Corresponding author.

*Abstract*— This essay is based on the joint project between the City University of Hong Kong and the 111, Inc. The pharmacy e-Commerce business grows rapidly in recent years with the ever-increasing medical demand during the pandemic. A big challenge for online pharmacy platforms is drug product matching. The e-Commerce platform usually collects their drug product information from multiple data sources such as the warehouse or retailers. Therefore, the data format is inconsistent, making it hard to identify and match the same drug product. This paper creates an integrated system for matching drug products from two data sources. Besides, the system would correct some inconsistent drug approval numbers based on a Naïve-Bayes drug type classifier. The matching system is a rule-based label prediction model, and the drug approval number correction system's main body is a Naïve-Bayes drug type classifier. Our integrated system achieves 98.3% drug matching accuracy, with 99.2% precision and 97.5% recall.

## I. Introduction

The Covid-19 pandemic significantly reduces outdoor activity and increase the demand for the medical product at the same time. Both of them facilitate the fast-growing online pharmacy industry, especially in a country with mature logistic and network infrastructure [1]. However, the drug products from those online platforms are not coming from a single source due to the company's complex store and supply system. The company's drug product information is usually collected from different sources, therefore in various data formats. The data format inconsistency brings difficulty in matching drug products coming from different sources [2]. This project uses field drug data coming from two data sources. For the product from each source, the drug name, drug manufacturer, approval number (like an ID for a drug product), and dosage information are given.

The barriers of drug product matching mainly from three aspects. The first is drug name format inconsistency, such as additive word, missing word, or unrelated information (mostly symbols). The second barrier for drug matching is due to the drug dosage format inconsistency. Some dosage descriptions from source one given in grams, but source two given in milligrams, and redundant and information existed in some rows. The last barrier is the result of the abnormal approval number. According to the China Food and Drug Administration rule, the approval number should be unique for each type of drug. However, when we investigate the dataset, some drug products are in the same label but different given approval numbers.

For drug name extraction, In 2011, Lee Peters [3] built a drug name retrieval system mainly based on the existing library. Lee's drug name extraction system starts with token splitting, followed by drug name similarity calculation (based on the intersection string). And finally, drug name extraction by linking it to the RxNorm library. Their method achieves 84%-92% matching accuracy in multiple datasets. However, our data drug name is given in Chinese and does not a normalized drug naming system, and the RxNorm is not applicable in the Chinese language. In 2018, L. Wang [4] built a drug parsing model starting from cleaning, semantic similarity comparison, and drug type recognition based on classical Chinese medicine dictionaries. Their model has 86% accuracy for matching Chinese Drugs. In our study, we develop a system that contains three sub-systems displayed in Fig.1.

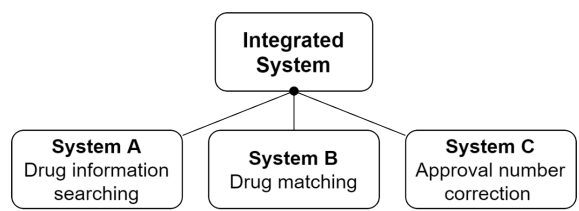

Fig. 1. The integrated system and three sub-systems

This paper consists of the following parts: Part II introduces the dataset and challenges. Part III illustrates the function and design logic behind the three sub-systems. Then, part IV concludes the drug matching accuracy with error analysis. And finally, part V gives the discussion and summarizes the result for this project.

## II. Dataset

Our dataset contains 29792 drug products retrieved from two data sources, and each data source has four attributes, added with the label column (label=1: match, label=0: not match). There is a total of nine columns of data. And for each attribute (an example row is shown in Table I below):

TABLE I.  AN EXAMPLE OF LABEL=1 ROW

| Attributes for a row of label=1 ||||
|---|---|---|---|
| *name_1* | *dosage_1* | *Manufactuer_1* | *approval_number_1* |
| 联环 叶酸片 | 4g（acid）*31 tablet | 江苏联环药业公司 | H20044917 |
| *name_2* | *dosage_2* | *Manufactuer_2* | *approval_number_2* |
| 叶酸片(盒) | 400mg*31 tablet | 江苏联环药业股份有限公司 | H20044917 |

Below are the problems about each attribute:

- Name: Extra unrelated words, symbols such as bracket make the matching not that intuitive.
- Dosage: the unit of measurement could be different. Some numbers are not providing helpful information about dosage (such as content in bracket).
- Manufacturer: strings are not precisely the same, and the drug product is not uniquely correlated with the manufacturer.
- Approval number: An ID number for drug products issued by China Food and Drug Administration (CFDA). The approval number has a fixed format: it starts with an alphabet that indicates its drug type, followed by 8-digit series. For example, Z11020204 begins with Z, which means the drug type is a traditional Chinese drug. The following table gives an example.

Besides, there are 156 abnormal approval number data in this dataset, which means the same drug product with two different approval numbers. This mainly results from the poor input data quality, and most of them can only be corrected manually. Therefore, the approval number cannot be used as hard evidence for the label due to those abnormal data points.

## III. System design and performance

The integrated drug product matching system comprises three sub-systems labeled as A, B, and C.

### A. Drug information searching system

Multiple manufacturers could produce a specific type of drug, and a particular manufacturer could make various types of drugs. Therefore, we built a system that allows users to input a drug name and get its popularity (appear times) and its manufacturer. The first step is to clean the drug name through the designed NLP process (remove symbols, remove the brand name, only keep drug product name). After that, we map the extracted drug name into correlated manufacturers. We use the fuzz ratio package [5] in Python, which is based on Levenshtein distance (the minimal single-character edits required to convert one string to another). The higher the fuzz ratio, the more similar the two strings. Therefore, the system able to detect the same manufacturers with tiny text variations.

For instance: the fuzz ratio of 悦康药业股份 and 悦康药业集团 pair is 93. Therefore, the system would recognize this pair as the same company. From this method, we could detect the duplicated company by looking at the fuzz ratio of n choose two combinations for drug companies (manufacturers) which more significant than 90. Then, export the detected duplicate and unique company lists. The whole system flow is shown in Fig 2.

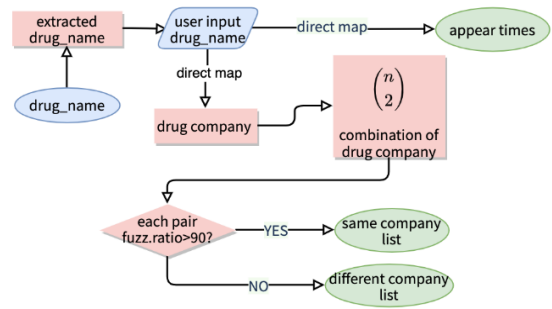

Fig. 2. System A: Drug information search system flow chart

And Table II below demonstrates the partial function of the drug information searching system (here, we didn't display the company list due to the space limitation).

TABLE II.  DEMONSTRATION FOR DRUG INFORMATION SYSTEM

| *Drug name* | *Duplicated manufacturers* | *Appear time* | *How many manufacturers* |
|---|---|---|---|
| 阿莫西林 | 悦康药业股份 悦康药业集团 | 257 times | 61 |

### B. Drug matching system (label prediction)

The drug matching system predicts the label for each row. It is determined by the dosage and name column of both data sources. The dosage column information could be split into two main parts: the dosage-related information and the package quantity related value. Table III shows an example of dosage data.

TABLE III.  AN INSTANCE OF DOSAGE COLUMN DATA STRUCTURE

| *Dosage_1* | *Dosage information* | *Package quantity* | *Label* |
|---|---|---|---|
| 0.3g*12 粒*2 板 | 0.3g | 12 粒*2 板 | |
| *Dosage_2* | *Dosage information* | *Package quantity* | 1 |
| 300mg*24s | 300mg | 24s | |

It can be clearly seen that the dosage data have a pretty regular format; hence with some cleaning, token splitting, we can extract the dosage and package quantity information for each row. So, the logic behind the whole matching system design is three main rules listed below:

1. The same drug must have the same dosage information.
2. The same drug must have the same package quantity.
3. The same drug must have a similar drug name string.

However, our dataset contains lots of points missing the dosage information but still label as 1. For instance: 0.35g*45s and 45s. If the dosage is missing, but the other two conditions meet, we still label it as 1. There are two reasons for this assumption: first, in our dataset, more than 80% of dosage missing cases are labeled as 1. And according to Gautam's study [6], the drug in the same type normally has fixed dosage/strength (based on domain knowledge and lab results).

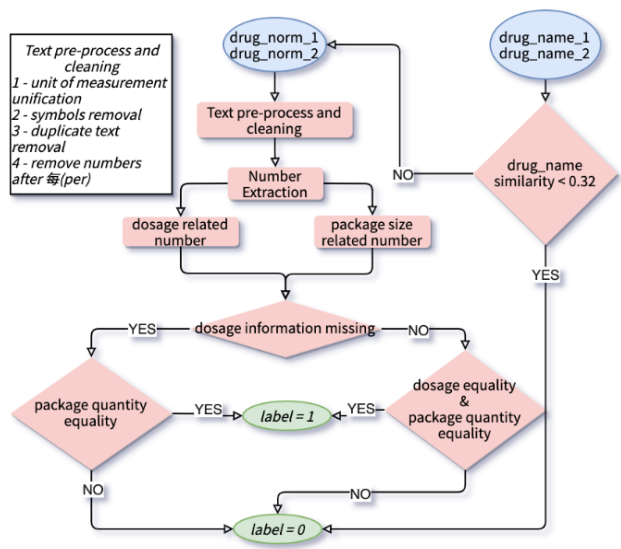

Fig. 3. System B: Drug matching system flow chart

According to those reasons, we made this reasonable assumption: when dosage information is missing and everything else is the same, we turn to the package size for label prediction. And based on the above logic, we design a rule-based label prediction system, and the process flow is shown in Fig.3. It turns out that our rule-based drug matching system performed well in this dataset (29792 data points) with accuracy =97.3%, precision =99.3%, and recall=97.5%.

*C. Inconsistent approval number correction system*

After label prediction, we could spot some inconsistent approval numbers with label=1 (odd approval number we mentioned in Part II: Dataset). The approval number is composed of:

- Starting alphabet reflects drug type: such as H, Z
- Eight digits after the alphabet: such as 20014542

So, the odd approval number could come from two parts. The inconsistency in digits is a random error and can only be corrected manually. However, the starting alphabet (reflects drug type) inconsistency could be rectified by the model. There are two types of drugs in our dataset, traditional Chinese drug (Z alphabet) and west drug (Non-Z alphabet). As can be seen in Fig 4, the traditional Chinese drug (Z drug) and west drug (NZ drug) naming styles are pretty different (few intersections for the high-frequency name forming a word).

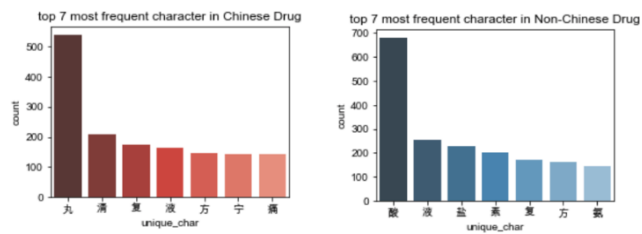

Fig. 4. The most frequent words in Chinese and non-Chinese drug name

According to W.cui [7], the multinomial Naïve-Bayes classifier suitable for dealing with text classification tasks, especially for the text of clear difference in the token frequency distribution. So, we build a Naïve-Bayes model to classify the Chinese drug (1=Traditional Chinese Drug, 0=Wes Drug) only based on their drug name. We first tokenize the drug name for each row and transform it to a matrix X. The matrix X has the number of features (1275) equals to the amount of unique word, and each row records its appearance count (5085 rows). The total data point is 5085. We choose a 10% test size, which achieves 97% accuracy. Fig.5 is the confusion matrix for the test data (precision = 0.97, Recall=0.96).

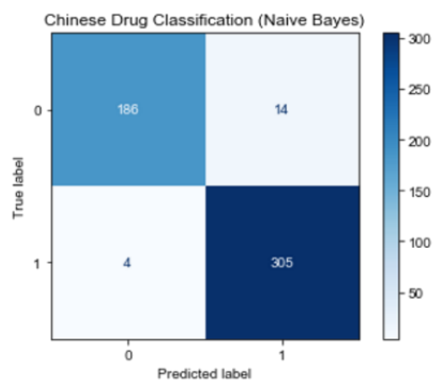

Fig. 5. The Naïve-Bayes Chinese drug classifier confusion matrix

Therefore, we could correct the Z-NZ type (one starts with Z, another is a non-Z start) approval number inconsistency. For example, in Table IV, we don't know which approval number is the correct one. Still, the traditional Chinese drug classifier suggests this given drug name is a traditional Chinese drug, so we should use the approval_number_1 that starts with Z., And we confirm our prediction via an online search.

TABLE IV. AN ILLUSTRATION OF APPROVAL NUMBER INCONSISTENCY

| name_1 =name_2 | Approval_ number_1 | Approval_ number_2 | label |
|---|---|---|---|
| 复方丹参片 | Z20173008 | H20046611 | 1 |

After label prediction via system B, we could check and auto-correct the Z-NZ type approval number consistency by system C. The process flow of system C is given in Fig.6.

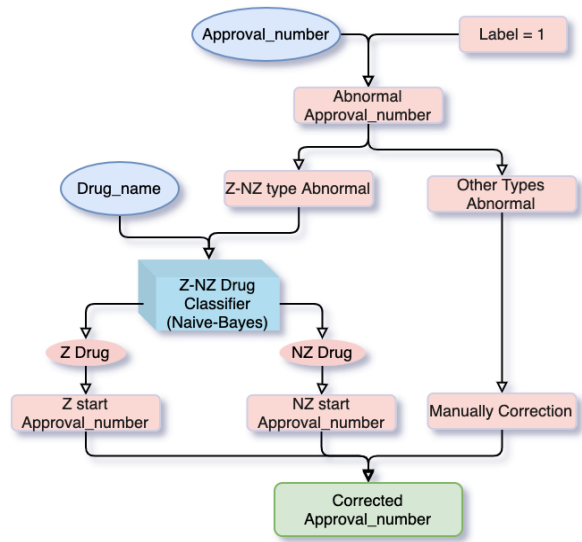

Fig. 6. System C: Inconsistent approval number correction system

## IV. DRUG MATCHING RESULTS

The principal object for this project is matching the drug product data coming from multiple data sources. Our matching system(System-B) shows 97.2% accuracy in this dataset. And we conduct an error analysis for our model prediction (shown in Fig 7). The error is mainly from the bad quality of data input in dosage and drug name, which is inevitable in an ample amount of text data. Besides, we also notice that 275 data are labeled wrong (the given label is wrong). After we corrected those wrong label points, we recalculate our factual drug matching accuracy, which rose to 98.3%.

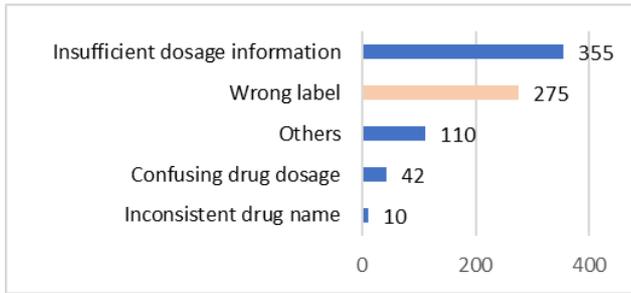

Fig. 7. Drug matching error analysis

## V. DISCUSSION AND CONCLUSION

In this project, we build a drug product matching system that successfully matches 29792 drug product data from two data sources with final accuracy equals 98.3%. Also, the user could get the popularity and manufacturer information for a specific drug. In System C, we build a Chinese drug classifier to auto-correct Z-NZ type abnormal approval number. This Naïve-Bayes drug name-based Chinese drug classifier achieves 97% accuracy. Fig 8 below shows our classifier's f-measure comparison to another dictionary-based Chinese drug recognition system built by L.Wang [4] in 2018.

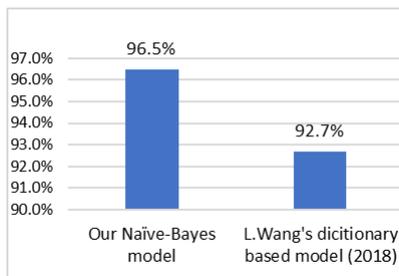

Fig. 8. Chinese drug classification F-measure comparison

Our integrated system composed of three sub-systems:

- System A - Drug information search system: this system will return the user input drug popularity(appear times) and its manufacturer's list.

- System B – Drug matching system: this is the main body of the whole design, which uses a rule-based method to predict the matching label for each row of data (98.3% accuracy)

- System C – Inconsistent approval number correction system: this system corrects the abnormal approval number involved with Z-NZ type (Chinese and Non-Chinese drug approval number inconsistency) built on a 97% accuracy Naïve-Bayes Chinese drug classifier. This classifier only takes the drug name as input.

In which System A could be applied separately just for drug information retrieving. Besides, System B and C are closely connected. In real life case, the data will first get the predicted label from System B, and then, the label=1 data point flows into System C for abnormal approval number detection and correction (as shown in Fig 9).

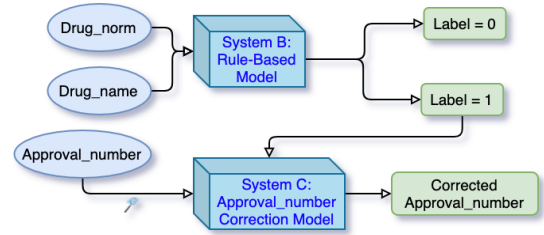

Fig. 9. The structure of connected System B and C

## ACKNOWLEDGMENT

I would like to express my gratitude to Professor Zhang Qingpeng, who provided me with valuable suggestions and patient guidance. His professional attitude and enthusiasm in medical research inspire me a lot. And The data is provided by 111, Inc.